\documentclass[11pt,authoryear]{article}

\usepackage{algorithm}
\usepackage{algorithmic}
\usepackage{natbib}
\usepackage{stfloats}
\usepackage{hyperref}
\usepackage{eurosym}
\usepackage{graphicx}
\usepackage{float}
\usepackage{booktabs}
\usepackage{footnote}	
\usepackage{wrapfig}
\usepackage{subcaption}
\usepackage{caption}
\usepackage[section]{placeins}
\usepackage{multirow}
\usepackage{multicol}
\usepackage{authblk}

\title{Towards automated feature engineering for credit card fraud detection using multi-perspective HMMs}

\date{\vspace{-5ex}}

\author[1,2]{Yvan Lucas}
\author[1]{Pierre-Edouard Portier}
\author[1]{L\'ea Laporte}
\author[3]{Liyun He-Guelton}
\author[3]{Olivier Caelen} 
\author[2]{Michael Granitzer}
\author[1]{Sylvie Calabretto}
\affil[1]{INSA Lyon}
\affil[2]{Universit{\"a}t Passau}
\affil[3]{Worldline Lyon}

\begin{document}


\maketitle

\begin{abstract}
Machine learning and data mining techniques have been used extensively in order to detect credit card frauds. However, most studies consider credit card transactions as isolated events and not as a sequence of transactions. 

In this framework, we model a sequence of credit card transactions from three different perspectives, namely (i) The sequence contains or doesn't contain a fraud (ii) The sequence is obtained by fixing the card-holder or the payment terminal (iii) It is a sequence of spent amount or of elapsed time between the current and previous transactions. Combinations of the three binary perspectives give eight sets of sequences from the (training) set of transactions. Each one of these sequences is modelled with a Hidden Markov Model (HMM). Each HMM associates a likelihood to a transaction given its sequence of previous transactions. These likelihoods are used as additional features in a Random Forest classifier for fraud detection. 

Our multiple perspectives HMM-based approach offers automated feature engineering to model temporal correlations so as to improve the effectiveness of the classification task and allows for an increase in the detection of fraudulent transactions when combined with the state of the art expert based feature engineering strategy for credit card fraud detection. 

In extension to previous works, we show that this approach goes beyond ecommerce transactions and provides a robust feature engineering over different datasets, hyperparameters and classifiers. Moreover, we compare strategies to deal with structural missing values.

\end{abstract}


\newpage
\section*{Introduction}

Credit card fraud detection presents several difficulties. One of them is the fact that the feature set describing a credit card transaction usually ignores detailed sequential information. Typical models only use raw transactional features, such as time, amount, merchant category, etc \cite{Donato1999}. \cite{Bolton2001} showed the necessity to use attributes describing the history of the transaction when they used unsupervised methods such as peer group analysis for credit card fraud detection. Consequently, \cite{Whitrow2008} create descriptive statistics as features in order to include historical knowledge. These descriptive features can be for example the number of transactions or the total amount spent by the card-holder in the past 24 hours for a given merchant category or country. \cite{Bahnsen2016} considered \cite{Whitrow2008} strategy to epitomized the state of the art feature engineering technique for credit card fraud detection. 

We identified several weaknesses in the construction of these features that motivated our work. First, descriptive statistics provide an aggregated view over a set of transactions. Such aggregated features do not consider fine-grained temporal dependencies between the transactions. For example, a common fraud pattern starts with low amount transactions for testing the card, followed by high amount transaction to empty the account. Second, aggregated features are usually calculated over transactions occuring in a fixed time window (e.g. 24 h). In general, transactions from very different card holders do not follow such a time pattern in general. However, the number of transactions made during such a time period can vary a lot for different card-holders. Fixed size aggregated statistics can't account for that fact. Third, these features consider only the history of the card-holder and do not exploit information of fraudulent transactions for feature engineering. However, a sequence of transactions happening at a fixed terminal can also contain valuable patterns for fraud detection.

In our work we propose to generate history-based features using Hidden Markov Models (HMM). They quantify the similarity between an observed sequence and the sequences of past fraudulent or genuine transactions observed for the cardholders or the terminals. 

We have chosen these three perspectives based on the following assumptions: the features made with only genuine historical transactions should model the common behavior of card holders. This is a classic anomaly detection scheme where the likelihood of a new sequence of  transactions is measured against historical honest sequences of transactions \cite{Chandola2012}. The other four features are made with sequences of transactions with at least one fraudulent transaction. The rationale is that to have a risk of fraud, it is not enough for a new sequence to be far from usual transaction behavior but it is also expected for it to be relatively close to a risky behavior. Thus, these last features should decrease the number of false positive. As stated by \cite{Pozzolo2017}, this is a crucial issue since the investigators can only verify a limited number of alerts each day. The second perspective allows the model to take the point of view of the card-holder and the merchant which are the two actors involved in the credit card transactions. The last perspective takes into account two important features for credit card fraud detection: the amount of a transaction and the elapsed time between two transactions. These features are strong indicators for fraud detection.

By using a \textbf{real world} credit card transactions dataset provided and labeled by a large European card processing company, we want to assess how much our contributions - namely the addition of the terminal perspectives (among others) in the construction of history features and the construction of multiple perspectives HMM-based features - improve fraud detection.

To quantify the impact of the addition of the HMM-based features, we use the Precision-Recall AUC metric. We fed Random Forest classifiers with transactional data including the state of the art transaction aggregation strategy and measured the increase in PR-AUC due to the HMM-based feature engineering technique we propose. The multiple perspective property of our HMM-based feature engineering allows for an incorporation of a broad spectrum of sequential information that leads to a significant increase of the detection of fraudulent transactions.

This paper consists in an extension to a poster \cite{Lucas2019}. As in \cite{Lucas2019}, we present the concept of multiperspective HMM-based feature engineering. Moreover, we extend  it significantly with additional experiments and evaluations. First of all, the framework is shown to increase the detection of fraudulent transactions not only for e-commerce transactions but also for face-to-face transactions. This result wasn't necessarily a foregone conclusion since e-commerce and face-to-face transactions present very different properties (e.g. merchant not open at night, necessity of a PIN authentication for face-to-face transactions...). Previous work done on the same dataset \cite{Jurgovsky2018} showed that some increase in detection observed on one type of transactions couldn't be extended to all types of transactions. Then, the feature engineering strategy is shown to be relevant for various types of classifiers (random forest, logistic regression and Adaboost) and robust to hyperparameters choices made for constructing the features: The number of hidden state and the length of the sequence have no strong effect on the quality of the fraud detection. Lastly, the framework suffered from a structural missing value limitation: For some users with few transactions, the HMM-based features couldn't be calculated. From 20\% to 40\% of the transactions (depending on the choice of the length of the sequences modeled by the HMM) are not associated with HMM-based features. After comparing several solutions to overcome this limitation, we were able to obtain a steady improvement of the detection over all the transactions of the dataset.

In this paper, we present the state-of-the-art approaches for the problems of credit card fraud detection and sequence classification in section \ref{related}. Afterwards, we show how the HMM-based features improve on the limitations of the state of the art techniques and the way they are created in section \ref{framework}. The experimental protocol is described in section \ref{expe}. In section \ref{results} and \ref{differentmodels}, we present the experimental results obtained on face-to-face and e-commerce transactions with different classifiers. The section \ref{params} is dedicated to prove the robustness of the method through an hyperparameter study. Finally, in section \ref{missing} we compare different solutions to tackle the issue of structural missing values.

The HMM-based features we propose present interesting assets in the context of credit card fraud detection and more generally anomaly detection. 

This work opens perspectives for feature engineering in any supervised task with sequential data. In order to ensure reproducibility, the source code of the proposed framework can be found at \url{https://gitlab.com/Yvan_Lucas/hmm-ccfd} .

\section{Related Work}
\label{related}
\subsection{\textbf{Credit card fraud detection}}
A wide range of machine learning approaches were used in credit card fraud detection. \citep{Maes2002} evaluated Artificial Neural Network and Bayesian belief network with ROC AUC on Europay International's dataset. 
\citep{Bhattacharyya2011} compared Support Vector Machine, Random Forest and logistic regression on a real world dataset using a wide variety of metrics. 
\citep{Bahnsen2013} adjusted Bayes Minimum Risk using real financial costs in order to adapt the prediction of Random Forest and Linear Regression classifier.
\citep{Pozzolo2014} tested an architecture to take into account temporal concept drift in the credit card transaction data stream with Random Forest, Support Vector Machine and Neural Network.
\citep{Mahmoudi2015} applied a modified Fisher discriminant function to take into account the higher false negative cost in credit card fraud detection. More recently, \citep{Jurgovsky2018} used LSTM for sequence classification on the same real world dataset that we use in this article. They showed that, in the case of face-to-face transactions only, sequence modelling with Long Short Term Memory networks (LSTM) improves fraud detection when compared to Random Forest with aggregated features.
	
Since Random Forest have been shown to perform well for credit card fraud detection in the litterature \citep{Bhattacharyya2011} and in preliminary experiments we have done, we chose to consider them in this work for evaluating the impact of our proposed features on the prediction quality. Moreover, Random Forests offer the possibility to calculate the importance of a feature which is defined as the decrease of gini impurity through a node weighted by the proportion of elements of the dataset passing through this node \citep{breiman1984}. This property is interesting for studying the impact of a feature engineering strategy.

Feature engineering is critical for credit card fraud detection. Some authors used only raw features in order to detect fraudulent transactions (\citep{Mahmoudi2015}, \citep{Minegishi2011}). \citep{Bolton2001} showed the necessity to use attributes describing the history of the transactions for unsupervised credit card fraud detection (peer group analysis). Lately, \cite{Saia2019} used Fourier and wavelet transforms in order to move the transaction in a new domain before applying a machine learning algorithm. This allows to raise outliers based on a different view of the dataset (frequential view). This is related to our approach since our approach consists in creating likelihood score for a variety of views on the dataset (sequential views).

\citep{Whitrow2008} proposed a transaction aggregation strategy to create descriptive features containing information about the past behaviour of the card-holder over a certain period of time. These descriptive features can be for example the number of transactions or the total amount spent by the card-holder in the past 24 h with the same merchant category or country. They showed a 28\% increase in the detection of fraudulent transactions by using these aggregated features with a Random Forest as the learning algorithm. 
\citep{Jha2012} applied Whitrow's transaction aggregation strategy to logistic regression in a real world credit card transactions dataset. 
\citep{Krivko2010} presented a rule-based approach using the difference between the recent amount spent by the card holder and either the average amount spent by this card holder or the average amount spent by all the card holders.  
\citep{Bahnsen2016} showed that adding periodic features based on the time of the transaction to Whitrow's aggregated features increases the savings by an average of 13\% with Random Forest, Logistic Regression and Bayes minimum Risk model.

An other feature engineering strategy has been proposed by \citep{Vlasselaer2015}. It consists in using the numbers of transactions between card holders and terminals in the graph of the transactions in order to create a time-dependent suspiciousness score. They showed a 3.4\% increase of the ROC AUC by using these network-based features together with Whitrow's aggregated features and Random Forests.

In this work, similar to other contributors, we consider Whitrow's card-holder centric aggregated features as the state of the art baseline for comparison with the HMM-based features we propose. 
\subsection{\textbf{Sequence Classification}}

Sequence classification is one of the main machine-learning research field. It aims to consider the sequential properties of the data at the algorithmic level in order to improve the classification of sequential data.
\cite{Dietterich2002} reviewed sequential classification based on sliding windows (or recurrent sliding windows). However, sliding windows methods don't take into account inner dependencies between consecutive events.

\cite{Srivastava2008} tried to overcome this limitation by using generative models such as Hidden Markov Models (HMMs) for credit card fraud detection. They motivated the choice of HMMs by relating the hidden states to the different types of purchase. For this purpose, they created an artificial credit card transactions dataset. In their multinomial HMMs, the transactions were characterized with a symbol ('big amount', 'medium amount', 'small amount') used as the observed variable. After training, the likelihood that the sequence of recent transactions was generated by the HMMs is computed. The decision is taken by comparing the likelihood to a threshold value.

\cite{Graves2012} observed that Long Short Term Memory networks are better than other sequential algorithms such as HMMs for speech recognition and handwriting recognition tasks since they allow to learn long term dependencies in sequences. \cite{Wiese2009} compared feed forward neural network combined with Support Vector Machines and LSTM for credit card fraud detection and showed that LSTM are relevant in the credit card fraud detection context because they can model time series of different length for each card holder. However, \cite{Jurgovsky2018} showed recently on the real world dataset that we use that LSTM offers a small improvemnt over Random Forest in the case of face-to-face transactions. There is no improvement for e-commerce transactions.

\section{Multiple perspectives Hidden Markov Model based feature engineering}

\label{framework}
The state of the art feature engineering techniques for credit card fraud detection creates descriptive features using the history of the card-holder (such as: "amount spent by the card-holder in shops from a given country in the last 24h" (\cite{Whitrow2008}, \cite{Bahnsen2016}). These descriptive features present several limits we aim to overcome. First, they do not take into account the history of the seller even if it is clearly identified in most credit card transactions dataset. Moreover these descriptive features do not consider dependencies between transactions of a same sequence. Therefore we use Hidden Markov Models which are generative probabilistic models and a common choice for sequence modelling \cite{Rabiner1991}. Finally, the choice of the descriptive feature created using the transaction aggregation strategy (\cite{Whitrow2008}, \cite{Bahnsen2016}) is guided by expert knowledge. In order not to depend on expert knowledge, we favor automated feature engineering in a supervised context.

\begin{figure}[h]
\hspace{-1.2cm}
\includegraphics[width=1.2\textwidth]{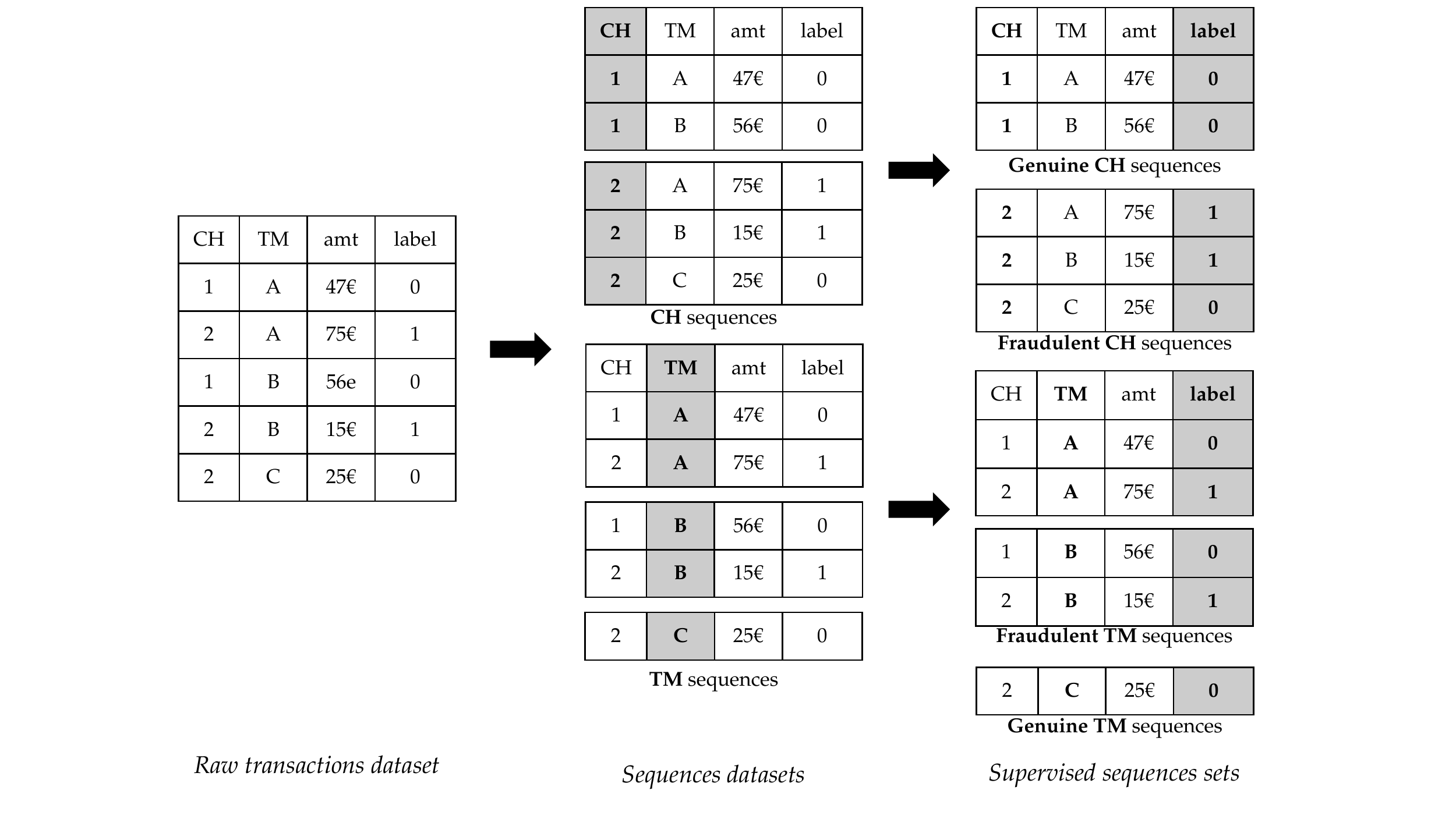}
\caption{Supervised selection of sequences for the training sets of the multiple perspectives Hidden Markov Models \textit{(CH$=$Card-holder, TM$=$Terminal)}}
\label{fig_framework}
\end{figure}
In addition to the descriptive aggregated features created by \cite{Whitrow2008}, we propose to create eight new HMM-based features. They quantify the similarity between the history of a transaction and eight distributions learned previously on set of sequences selected in a supervised way in order to model different perspectives.

We model the sequence of transactions from the combinations of three binary perspectives (genuine / fraudulent, card-holder / merchant, amount / timing) and therefore learn eight ($2^3$) different HMMs. At the end the set of 8 HMM-based features will provide information about the genuineness and the fraudulence of both terminal and card holder histories.

In particular, we select three perspectives for modelling a sequence of transactions (see figure \ref{fig_framework}). A sequence (i) can be made only of genuine historical transactions or can include at least one fraudulent transaction in the history, (ii) can come from a fixed card-holder or from a fixed terminal, and (iii) can consist of amount values or of time-delta values (i.e. the difference in time between the current transaction and the previous one). We optimized the parameters of eight HMMs using all eight possible combinations (i-iii). 


To learn the HMM parameters on observed data, we create 4 datasets:
\begin{itemize}
\item sequences of transactions from \textbf{genuine credit cards} (without fraudulent transactions in their history).
\item sequences of transactions from \textbf{compromised credit cards} (with at least one fraudulent transaction)
\item sequences of transactions from \textbf{genuine terminals} (without fraudulent transactions in their history)
\item sequences of transactions from \textbf{compromised terminals} (with at least one fraudulent transaction)
\end{itemize}

We then extract from these sequences of transactions the symbols that will be the observed variable for the HMMs. In our experiments, the observed variable can be either:
\begin{itemize}
\item the amount of a transaction.
\item the amount of time elapsed between two consecutive transactions of a card-holder (time-delta).
\end{itemize}

%

\begin{figure}[H]
\centering
\includegraphics[scale=0.35]{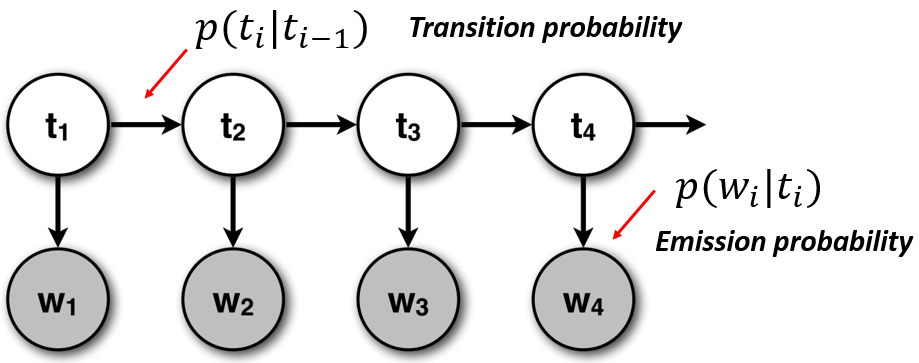}
\caption{Hidden markov model architecture.}
\label{fig:HMM}
\end{figure}

Hidden Markov Models main hypothesis is that behind the observed distribution, there is a simpler latent model that rules the sequential distribution. Hypothetically, the complexity of the observed distribution comes partly from additive Gaussian noise corrupting both the state evolution process and the emission process. Hidden Markov Models allow for a compression of the observed sequence in order to generalize the observed behaviour into an abstracted latent behaviour. It is comprised of two processes represented by matrices (see figure \ref{fig:HMM}):
\begin{itemize}
\item The \textit{transition matrix} describes the evolution of the hidden states. Each row $i$ of the transition matrix is a multinomial distribution of the next state given that the current state is $i$. The hidden states obey to the Markov property (i.e. given the present the future does not depend on the past).
\item The \textit{emission matrix} describes the conditional distribution of the observed variables given the current hidden state. Usually, the distribution is considered multinomial for categorical observed variables or gaussian for continuous observed variables.
\end{itemize}

The transition and emission conditional probability matrices of the HMMs are optimized by an iterative Expectation-Maximisation algorithm known as the Baum-Welch algorithm (\cite{Baum1972}, \cite{Rabiner1991}).
EM optimization of a model with latent (hidden) parameters consists in (for parameters initalized with a value):
\begin{description}
\item[Expectation:] Find the latent states distributions that correspond the most to the sequences of observed data. This is usually done with the help of the Viterbi algorithm which recursively leverage the Markov property in order to simplify the calculations of the conditional probabilities to observe a sequence of event given the parameters of the transition and emission matrices (also referred to as forward-backward algorithm \cite{viterbi1967}).
\item[Maximisation:] Maximise the correspondence between the latent distribution inferred during the expectation step and the parameters of the transition and emission matrices by adjusting the parameters.
\end{description}
The expectation and maximization steps are repeated until convergence of the graphical model to the observed data. The convergence can be monitored by observing the increase of the value of the likelihood that the set of observed sequences has been generated by the model. This likelihood increases over the iterations until it reaches a ceiling when the hyper parameters ruling the architecture of the generative model don't allow it to fit more to the set of observed sequences.


At the end, we obtain 8 trained HMMs modeling 4 types of behaviour (genuine terminal behaviour, fraudulent terminal behaviour, genuine card-holder behaviour and fraudulent card-holder behaviour) for both observed variables (amount and time-delta).

\begin{algorithm}[h]
\caption{Online: calculate likelihood of sequences of observed events}
\begin{algorithmic}
\FOR {$tx_{i}$ in all transactions}
\FOR {$perspective_{j}$ in perspectives combinations}
\STATE $[tx_{i}, tx_{i-1}, tx_{i-2}] \leftarrow  user sequence_{i,j}$
\STATE $HMM_{j} \leftarrow HMM(usertype, signaltype, sequencetype)$
\STATE $AnomalyScore \leftarrow  log(P([tx_{i},tx_{i-1},tx_{i-2}]|HMM_{j}))$
\ENDFOR
\ENDFOR
\end{algorithmic}
\end{algorithm} 	

The HMM-based features proposed in this paper (table \ref{HMMfeats}) are the likelihoods that a sequence made of the current transaction and the two previous ones from this terminal/card holder is generated by each of these models. In order to calculate their value, the most probable sequence of hidden states for each observed sequence has to be computed. This is usually done with the help of the Viterbi algorithm \cite{viterbi1967} which also leverages the Markov property in order to simplify the calculations of the conditional probabilities to observe a sequence of hidden states given the parameters of the HMM (initial probabilities and transition and emission matrices) and an observed sequence of events.

\begin{table}[h]
\centering
\begin{tabular}{cc|cc}
User & Feature & Genuine & Fraudulent\\
\hline
\multirow{2}{*}{Card Holder}& Amount & \textit{HMM1} & \textit{HMM5} \\
& Tdelta & \textit{HMM2} & \textit{HMM6}\\
\multirow{2}{*}{Terminal} & Amount & \textit{HMM3} & \textit{HMM7}\\
& Tdelta & \textit{HMM4} & \textit{HMM8}\\
\end{tabular}
\caption{Set of 8 HMM-based features describing 8 combinations of perspectives \label{HMMfeats}}
\end{table}

\section{Experimental Setup}
\label{expe}
\subsection{\textbf{Dataset Description}}
%

We used a credit card transactions dataset provided by our industrial partner in order to quantify the increase in detection when adding HMM-based features. This dataset contains the anonymized transactions from all the belgian credit cards between 01.03.2015 and 31.05.2015.

The classification task is to predict the class of the transactions (genuine or fraudulent).

Transactions are represented by vectors of continuous, categorical and binary features that characterize the card-holder, the transaction and the terminal. The card-holder is characterized by a unique card-holder ID, its age, its gender, etc. The transaction is characterized by variables like the date-time, the amount and other confidential features. The terminal is characterized by a unique terminal-ID, a merchant category code, a country.

In addition to the work already presented as a poster \cite{Lucas2019}, we studied the impact of the proposed multiple perspective HMM-based feature engineering strategy for face-to-face transactions which present very different properties than e-commerce transaction: The merchant is usually closed at night, the necessity of a PIN authentication decreases significantly the number of fraudulent transactions \cite{Ali2019}, etc. For comparison, \cite{Jurgovsky2018} have shown on the same belgium transactions dataset that some machine learning approaches (LSTM) gave better results on face-to-face transactions than on e-commerce transactions.

\subsection{\textbf{Feature Engineering and Dataset Partitioning}}


%

In order for the HMM-based features and the aggregated features to be comparable, we calculate terminal-centered aggregated features in addition to \citep{Whitrow2008} card-holder centered aggregated features (table \ref{aggTM})

\begin{table}[H]
\begin{tabular}{l|l}  
Feature & Signification\\
\hline
AGGCH1 & $\#$ transactions issued by user in 24h.\\
AGGCH2 & Amount spent by user in 24h.\\
AGGCH3 & $\#$ transactions in the country in 24h.\\
AGGCH4 & Amount spent in the country in  24h.\\
\hline
AGGTM1 & $\#$ transactions in terminal in 24h.\\
AGGTM2 & Amount spent in terminal in 24h.\\
AGGTM3 & $\#$ transactions with this card type in 24h.\\
AGGTM4 & Amount spent with this card type in 24h.\\
\end{tabular}

\caption{Aggregated features centered on the card holders and the terminal}
\label{aggTM}
\end{table} 	


We split temporally the dataset in three different parts: the training set, the validation set and the testing set. We chose to separate the time period corresponding to the training and validation set and the time period corresponding to the testing set with a gap of 7 days. The reason is that for real world fraud detection systems, human investigators have to verify the alerts generated by the classifiers. Since this process takes time, the ground truth is delayed by about one week. The transactions appearing during this gap of 7 days before the testing set are used in order to calculate the value of the aggregated and HMM-based features in the testing set but not for training the classifiers. Therefore, the training set contains all the transactions between 01.03.2015 to 26.04.2015, the validation set used for the tuning of the random-forests contains all the transactions between 27.04.2015 to 30.04.2015 and the testing set contains all the transactions between 08.05.2015 to 31.05.2015.

\section{Results}
\subsection{Improvement in fraud detection when using HMM-based features}
\label{results}

\begin{figure}[h]
\centering{
\vspace{-0.3cm}
\includegraphics[width=\textwidth]{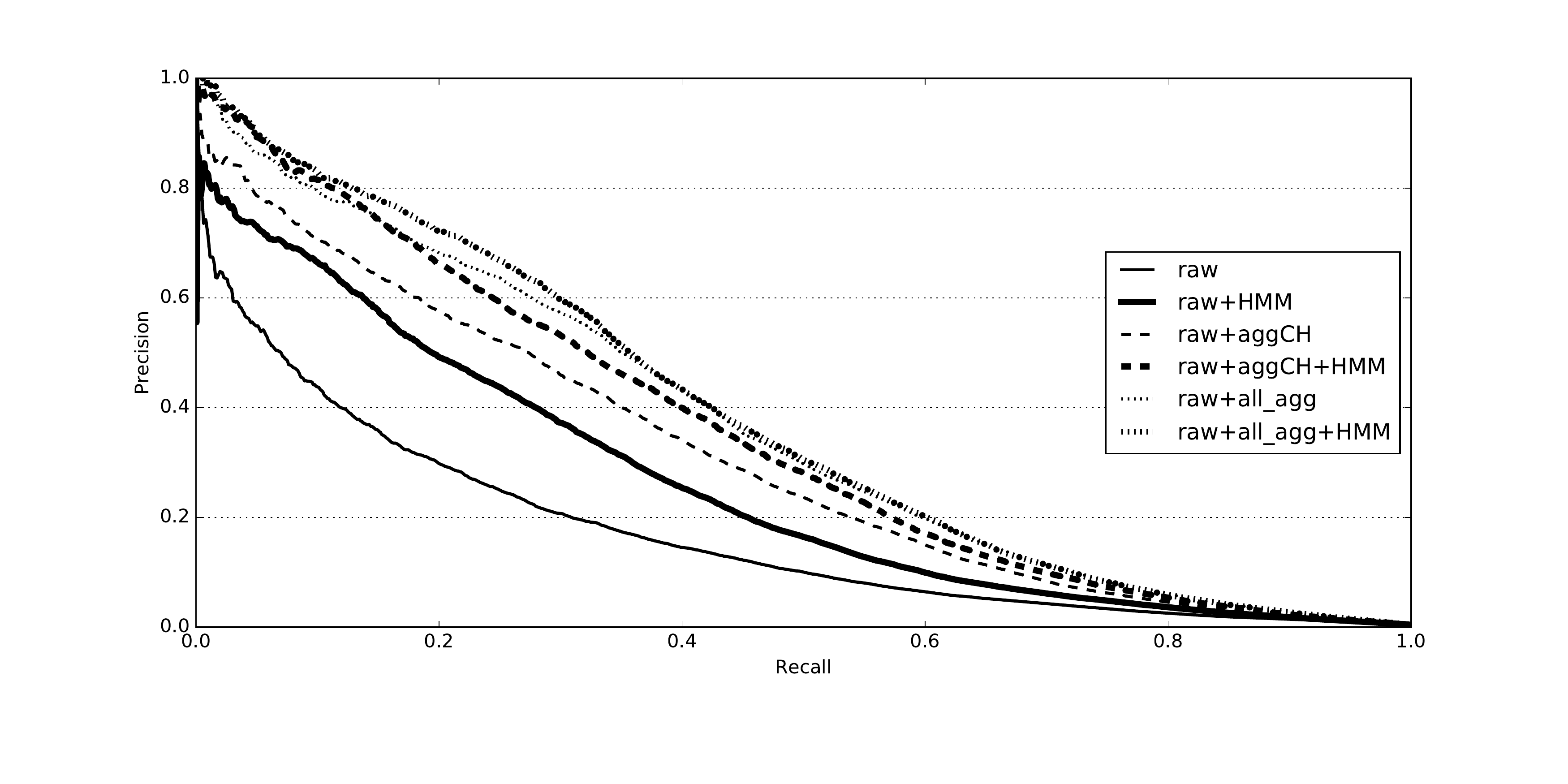}
\vspace{-0.2cm}
\begin{tabular}{l|ccc}  
Feature set & no HMM-features & HMM-features & increase through HMMs\\
\hline
raw & 0.212 $\pm$ 0.009 & 0.298 $\pm$ 0.0006 & 40.5\% \\
raw+aggCH & 0.343 $\pm$ 0.001 & 0.375 $\pm$ 0.0006 & 9.3\% \\
raw+all agg & 0.383 $\pm$ 0.002  & 0.397 $\pm$ 0.003 & 3.6\%\\
\hline
\end{tabular}
\caption{Precision-recall curves for e-commerce transactions. \textit{(Each color corresponds to a specific feature set, the line style corresponds to the presence or not of HMM-based features. The addition of HMM-based features (bold lines) to each feature set considered - even the most informative ones - allows for an increase in the detection of fraudulent transactions when compared to the same prediction without HMM-based features (thin lines).)}\label{ecomresults}
}}
\end{figure}

\begin{figure}[h]
\centering{
\vspace{-0.3cm}
\includegraphics[width=\textwidth]{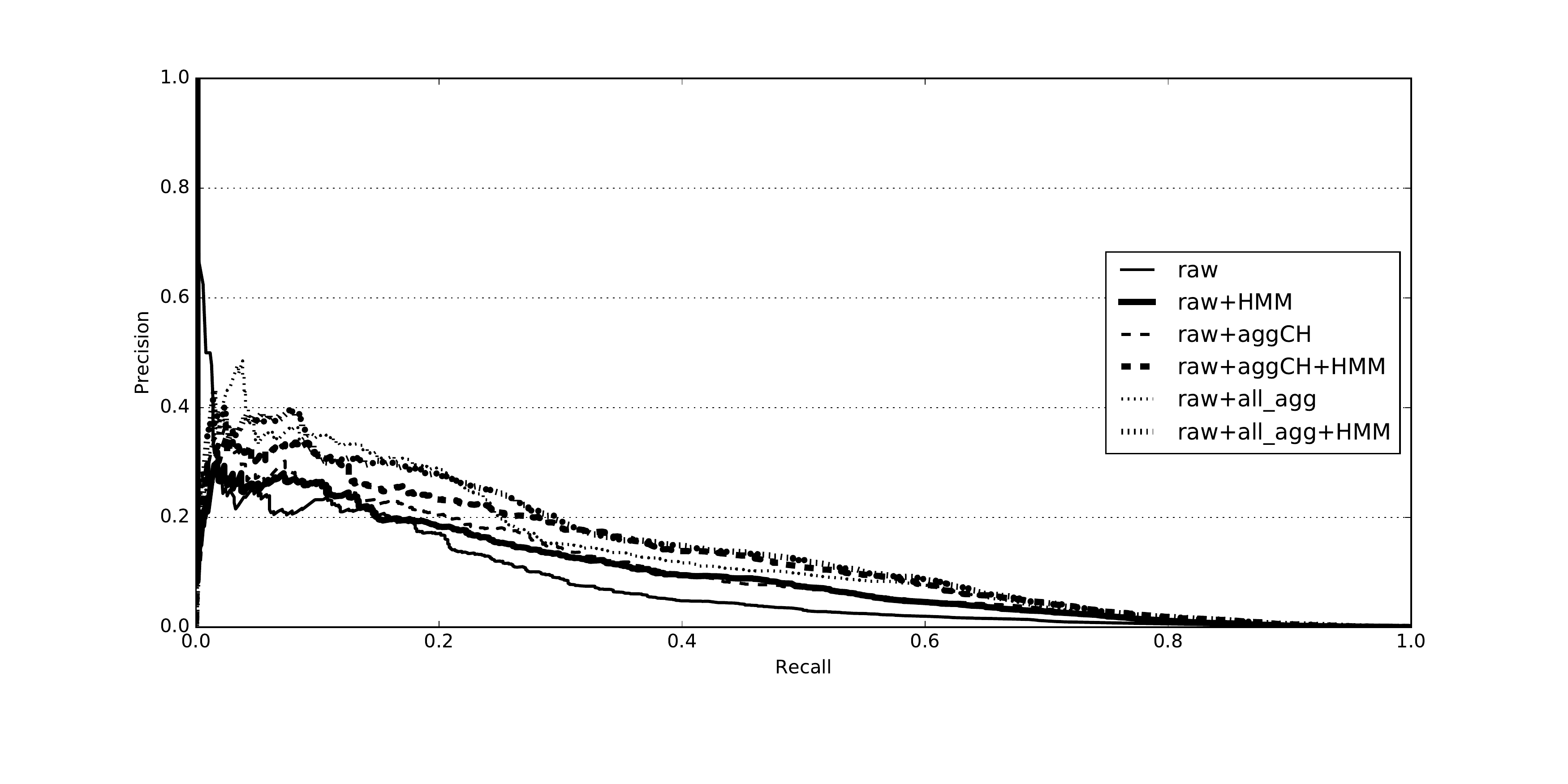}
\vspace{-0.2cm}
\begin{tabular}{l|ccc} 
Feature set & no HMM features & HMM features & increase through HMMs\\
\hline
raw & 0.082 $\pm$ 0.001 & 0.152 $\pm$ 0.001 & 85.4\% \\
raw+aggCH & 0.144 $\pm$ 0.001 & 0.170 $\pm$ 0.001 & 18.1\% \\
raw+all agg & 0.167 $\pm$ 0.001 & 0.177 $\pm$ 0.0006 & 6.0\%\\
\hline
\end{tabular}
\caption{Precision-recall curves for face-to-face transactions.\label{f2fresults}
}}
\end{figure}

We train Random Forest Classifiers using different feature sets in order to compare the efficiency of prediction when we add HMM-based features to the classification task for the face-to-face and e-commerce transactions. The difference in term of raw AUC between the e-commerce and the face-to-face transactions is due to the difference in imbalancy between these datasets: the imbalancy is 17 times stronger for the face-to-face transactions than for the e-commerce transactions. For face-to-face, there are 0.2 frauds per 1000 transactions whereas for e-commerce there are 3.7 frauds per 1000 transactions.

We tested the addition of our HMM-based features to several feature sets. We refer to the feature set \textbf{”raw+aggCH”} as the state of the art feature engineering strategy since it contains all the raw features with the addition of Whitrow’s aggregated features \cite{Whitrow2008}. The feature groups we refer to are: the raw features (raw), the features based on the aggregations of card-holders transactions (aggCH), the features based on the aggregation of terminal transactions (aggTM), the proposed HMM-based features (HMM features). 

In this section, the HMMs were created with 5 hidden states and the HMM-based features were calculated with a window-size of 3 (actual transaction + 2 past transactions of the card-holder and of the terminal). We showed in section \ref{params} that the HMM hyperparameters (number of hidden states and size of the window considered for the calculation of HMM-based features) did not change significantly the increase in Precision-Recall AUC observed.

We tuned the Random Forest hyperparameters (table \ref{gridRF}) through a grid search that optimizes the Precision-Recall Area under the Curve on the validation set. The choice of the Precision-Recall AUC for imbalanced dataset was motivated by the work of \cite{Davis2006}. 

\begin{table}[H] 
\centerline{
\begin{tabular}{c|c|c|c}
n-trees & n-features & min-samples-leaf & max depth\\
\hline
$\left\{300\right\}$ & $\left\{1, 7, 13\right\}$ & $\left\{1, 20, 40\right\}$ & $\left\{ 4, None\right\}$ \\ 
\end{tabular}}
\caption{Random Forest grid search
\label{gridRF}}
\end{table}

Figures \ref{ecomresults} and \ref{f2fresults} show the precision recall curves and their AUC obtained by testing the efficiency of Random Forests trained with several feature sets on the transactions of the testing set. The AUC numbers correspond to the average $\pm$ standard deviation on 3 different runs. The AUCs are stable over the different runs and the standard deviation numbers are low. 

We can observe that the face-to-face results (figure \ref{f2fresults}) and the e-commerce results (figure \ref{ecomresults}) both show a significant improvement in precision-recall AUC when adding sequence descriptors such as the proposed HMM-based features or Whitrow’s aggregated features to the raw feature set. For the e-commerce transactions, this improvement ranges from 3.6\% for the best feature set without HMM-based features (raw+all-agg for the e-commerce transactions) to 40.5\% for the worst feature set (raw for the e-commerce transactions). For the face-to-face transactions, this improvement ranges from 6.0\% for the best feature set without HMM-based features (raw+all-agg for the face-to-face transactions) to 85.4\% for the worst feature set (raw for the face-to-face transactions). 

By comparing the AUC of the curves raw+aggCH and raw+aggCH+HMM, we observe that adding HMM-based features to the state of the art feature engineering strategy introduced in the work of \cite{Whitrow2008} leads to an increase of 18.1\% of the PR-AUC for the face-to-face transactions and to an increase of 9.3\% of the PR-AUC for the e-commerce transactions. 

The relative increase in PR-AUC when adding terminal centered aggregated features to the feature set is of 16.0\% for the face-to-face dataset. The relative increase in PR-AUC when adding terminal centered aggregated features to the feature set is of 11.7\% for the e-commerce transactions. 

Overall, we can observe that the addition of features that describe the sequence of transactions, be it HMM-based features or Whitrow’s aggregated features, increases a lot the Precision-Recall AUC on both e-commerce and face-to-face transactions. The addition of HMM-based features improves the prediction on both e-commerce and face-to-face transactions and allows the classifiers to reach the best levels of accuracy on both e-commerce and face-to-face transactions.

\subsection{Robustness over different machine learning algorithms}
\label{differentmodels}
We have shown in section \ref{results} that the addition of HMM-based features to the existing transaction aggregation strategy allows for a consistent and significant increase in the precision-recall AUC of random forest classifiers.

In order to make sure that this increase is stable over different types of classifiers, we did the same experiments with adaboost and logistic regression classifier.

We tuned the gradient boosting and logistic regression hyperparameters (table \ref{gridmodels}) through a grid search that optimizes the Precision-Recall Area under the Curve on the validation set.

\begin{table}[h] 
\centerline{
\begin{tabular}{c|c|c|c}
tree numbers & learning rate & tolerance for stopping & max tree depth\\
\hline
$\left\{100, 400\right\}$ & $\left\{0.1, 1, 100\right\}$ & $\left\{10, 100\right\}$ & $\left\{1, 4\right\}$ \\ 
\end{tabular}}
\caption{Adaboost grid search}
\centerline{
\begin{tabular}{c|c|c}
C parameter & penalty & tolerance for stopping\\
\hline
$\left\{1, 10, 100\right\}$ & $\left\{l1, l2\right\}$ & $\left\{10, 100\right\}$ \\ 
\end{tabular}}
\caption{Logistic regression grid search
\label{gridmodels}}
\end{table}

\begin{figure}[h]
\centering{
\begin{tabular}{l|ccc}  
E-commerce  & no HMM-features & HMM-features & increase through HMMs\\
\hline

raw & 0.015 $\pm$ 0.0001 & 0.049 $\pm$ 0.001 & 267\% \\
raw+aggCH & 0.092 $\pm$ 0.005 & 0.121 $\pm$ 0.003 & 31.5\% \\
 raw+all agg & 0.096 $\pm$ 0.004  & 0.124 $\pm$ 0.005 & 29.2\%\\
 \hline
 \end{tabular}
  \begin{tabular}{l|ccc}
   Face-to-face  & no HMM-features & HMM-features & increase through HMMs\\ 	
\hline
 raw & 0.0094 $\pm$ 0.0001 & 0.030 $\pm$ 0.002 & 219\% \\
 raw+aggCH & 0.014 $\pm$ 0.001 & 0.0247 $\pm$ 0.02 & 76.4\% \\
 raw+all agg & 0.0161 $\pm$ 0.0003 & 0.0315 $\pm$ 0.0007 & 95.6\%\\
\hline
\end{tabular}
\caption{Precision-recall AUC for logistic regression classifiers.\label{logregAUCs}}
}
\end{figure}

\begin{figure}[h]
\centering{
\begin{tabular}{l|ccc}  
E-commerce  & no HMM-features & HMM-features & increase through HMMs\\
\hline
raw & 0.103 $\pm$ 0.001 & 0.163 $\pm$ 0.007 & 58.3\% \\
raw+aggCH & 0.238 $\pm$ 0.009 & 0.247 $\pm$ 0.011 & 3.8\% \\
raw+all agg & 0.263 $\pm$ 0.004  & 0.264 $\pm$ 0.006 & 0.4\% \\
 \hline
 \end{tabular}
 
  \begin{tabular}{l|ccc}
   Face-to-face  & no HMM-features & HMM-features & increase through HMMs\\ 	
\hline
raw & 0.034 $\pm$ 0.003 & 0.058 $\pm$ 0.009 & 70.6\% \\
raw+aggCH & 0.048 $\pm$ 0.002 & 0.079 $\pm$ 0.006 & 64.6\% \\
raw+all agg & 0.053 $\pm$ 0.003 & 0.061 $\pm$ 0.004 & 15.1\%\\
\hline
\end{tabular}
}
\caption{Precision-recall AUC for Adaboost classifiers.\label{adaAUCs}}
\end{figure}

By comparing the AUCs reported on table \ref{logregAUCs}  and \ref{adaAUCs}, we can conclude that the improvement observed when integrating the proposed multiple perspectives HMM-based feature engineering in addition to the state of the art transaction aggregation strategy is significant and reliable over different classifiers and datasets. The relative improvement is bigger for weaker classifiers such as logistic regression classifiers than for strong credit card fraud detection classifiers such as random forest classifiers.   

We observe a significant decrease in detection efficiency with logistic regression classifier compared to random forest classifier. Logistic regression classifier is known to be weak to categorical features encoded with a label encoder (category transformed to numbers) since the label encoding implies an order between the categories that is irrelevant. The results obtained with logistic regression classifiers may be improved using one-hot encoding or frequency encoding like advised in some other credit card fraud detection project (\cite{Pozzolo2015}, \cite{Jurgovsky2018}), however the evolution observed when including the proposed feature engineering strategy to logistic regression classifier is consistent with what we observed using random forest classifier.

We observed that shallow adaboost classifiers with a tree depth restricted to 1 as adviced in the litterature \cite{Freund1997} lead to significantly worse detection of fraudulent transactions than deeper adaboost classifiers using trees of depth 4. The downside of deeper trees is the increased risk of overfitting of the adaboost classifiers. Monitoring the number of iterations (tree added) after which the classifier starts to overfit could be done by tracking the PR-AUC on the validation set and on the training set. Due to adaboost design, the PR-AUC on the training set should increase for each tree added, however when the classifier overfits, the PR-AUC on the validation set would decrease: the classifier will start to learn particularities of the training set that are not relevant in other sets.


\subsection{Robustness to hyperparameters changes}
\label{params}

In order to understand if the feature engineering strategy is sensitive to the hyperparameters used for the construction of the HMM-based features, we constructed 9 sets of HMM-based features with different combinations of the HMM-based features hyperparameters.
The hyperparameters considered are the number of hidden states in the HMMs and the size of the sequence of past transactions used for the calculation of the likelihood score.

We measure the AUC obtained on the test set when adding different set of HMM-based features obtained with different combinations of hyperparameters to the raw feature set.

\begin{table}[h]
\centering
\begin{tabular}{lc|ccc}
& &\multicolumn{3}{c}{Window size}\\
 & & 3&5 &7 \\
\hline
Hidden& 3 &\multicolumn{1}{c}{0.280 $\pm$ 0.013} &\multicolumn{1}{c}{0.292 $\pm$ 0.015} &\multicolumn{1}{c}{0.295 $\pm$ 0.019}\\
states& 5 & \multicolumn{1}{c}{\textbf{0.315 $\pm$ 0.008}}&\multicolumn{1}{c}{0.310 $\pm$ 0.006} &\multicolumn{1}{c}{0.297 $\pm$ 0.007}\\
& 7 & \multicolumn{1}{c}{0.307 $\pm$ 0.003}&\multicolumn{1}{c}{0.307 $\pm$ 0.004} &\multicolumn{1}{c}{0.296 $\pm$ 0.011}\\
\end{tabular}
\caption{PR-AUCs for E-commerce HMM hyperparameters ($raw=0.203\pm0.005$)}
\end{table}

\begin{table}[h]
\centering
\begin{tabular}{lc|ccc}
& &\multicolumn{3}{c}{Window size}\\
& & 3&5 &7 \\
\hline
Hidden& 3 &0.159 $\pm$ 0.020&0.135 $\pm$ 0.007&0.150 $\pm$ 0.013\\
states& 5 & \textbf{0.139 $\pm$ 0.006}&0.135 $\pm$ 0.012&0.123 $\pm$ 0.009\\
& 7 & 0.129 $\pm$ 0.008&0.131 $\pm$ 0.009&0.124 $\pm$ 0.007\\
\end{tabular}
\caption{PR-AUCs for Face-to-face HMM hyperparameters ($raw=0.089\pm0.011$)}
\end{table}

We observe that, the combination of parameter $\{$window size: 3, hidden states: 5$\}$ gives the best AUCs on average on 3 runs. However the standard deviation values ($\pm$) are too high to confidently say that a hyperparameters combination is better than all the others.

A window size of 3 and 5 hidden states was the combination of hyperparameter chosen for the construction of the HMM-based features of the section \ref{results}.

In a real world case, doing the hyperparameter search can be worth it but requires to calculate several set of HMM-based features. Overall we can conclude that the HMM-based feature engineering strategy is relatively stable over the hyperparameters.

We also observed that adding all 9 newly calculated HMM-based features sets provided a slight improvement on the face-to-face transactions (13.3\%). We thought that adding HMM-based features with different hyperparameters could possibly bring different type of information. For example the window-size parameter could be seen as: description of short-term/mid-term/long-term history of the transaction. This small improvement wasn't observed on the e-commerce transactions (8.5\% decrease).

\section{Handling the missing value limitations}
\label{missing}

\begin{table}[h]
\begin{tabular}{l|cc|cc}
& \multicolumn{2}{c}{e-commerce}& \multicolumn{2}{|c}{face-to-face}\\
& \# transactions & \# frauds & \# transactions & \# frauds\\
\hline
\textit{All transactions} & $5.4*10^6$& $18472$ & $6.8*10^6$ & $2764$\\
$History>=3$& $4.5*10^6$&$15650$&$5.3*10^6$&$1305$\\
$History>=7$& $3.4*10^6$&$12493$& $4.1*10^6$&$859$\\
\end{tabular}
\caption{Number of transactions in the test set with different history constraints ($History >=3$ means that all the transactions have at least 2 transactions in the past for the card-holder \textbf{and} for the terminal: we can build sequences of 3 transactions for both perspectives)}
\label{ntxs}
\end{table}

With the HMM feature engineering framework, we can calculate sets of HMM features for different window sizes. However, when the transaction history is not big enough we can't calculate the HMM-based features for it. Because of this limitation we had to dismiss about 20\% of the transactions for the experiments described in section \ref{results} and around 40\% of the transactions for the experiments of the hyperparameter section \ref{params}. There is a strong need to tackle the issue of structural missing values depending on the length of users sequences in order to integrate transactions with short history where part or all of the HMM-based features couldn't have been calculated.

In this section we consider 9 sets of HMM based features obtained with a window size of 3, 5 and 7 for the card-holder or for the terminal. We therefore have 16 sets of transactions with different history constraints (terminal history: [0, 3, 5, 7] $*$ card-holder history: [0, 3, 5, 7]).

We consider 3 missing values strategies:
\begin{description}
\item[default0] a genuine default value solution where the HMM-based features that couldn't have been calculated for the current transactions are replaced by 0.
\item[weighted PR] a weighted sum of the predictions of the Random Forests specialized on the history constraints of the transaction, among 16 Random Forests each trained on one of the 16 possible history constraints. For example, for a transaction with a terminal history of 6 and a card-holder history of 3 we will sum the predictions of the Random Forests [0,0], [0, 3], [3, 0], [3, 3], [5, 0], [5, 3]\footnote{The first number describes the terminal history constraint, the second number describes the card-holder history constraint}. Each Random Forest is weighted by it's efficiency on the validation set. We used PR AUCs values as weights for the Random Forests.
\item[stacked RF] a stacking approach where a Random Forest classifier is trained on the predictions of the 16 Random Forests specialized on the constraints (0 for missing prediction, when the transaction doesn't satisfy the constraints of the considered Random Forest). This approach has the second benefit that we might stack indivual classifiers thereby creating a more accurate new one.
\end{description}
Other approaches to handle missing values advise to generate them by modelling the distribution of the corresponding features. We considered that these solutions don't apply in our case since the missing values appear because there wasn't enough historical information to calculate the corresponding feature. We thought that replacing the value of a model based feature (HMM based features) using an other model (kNN or Random Forest) wasn't an appropriate solution.

\begin{table}[h]
\begin{tabular}{l|c|c}
& \multicolumn{1}{c}{e-commerce}& \multicolumn{1}{|c}{face-to-face}\\
& PR-AUC &PR-AUC \\
\hline
\textit{raw (no HMM-based features)} & $0.264\pm0.002$& $0.058\pm0.012$\\
\textit{default0}& $0.320\pm0.001$&$0.145\pm0.004$\\
\textit{weighted PR}& $0.275\pm0$& $0.040\pm0.002$\\
\textit{stacked RF}& $0.317\pm0.002$& $0.133\pm0.021$\\
\end{tabular}
\caption{Fraud detection on the whole test set with different missing values strategies}
\label{results-missing}
\end{table}

We can observe in table \ref{results-missing} that the simplest solution (\textit{default0}) allows for the best PR-AUCs with the best stability for both face-to-face and e-commerce transactions. This is also by far the fastest method since it needs to only train one Random Forest instead of 16 (resp. 17) for the \textit{weighted PR} approach (resp. \textit{stacked RF} approach).

The \textit{weighted PR} solution doesn't allow for satisfying results. However the \textit{stacked RF} solution gives good PR-AUCs and presents interesting properties in order to combine different types of classifiers.

Finding satisfying solutions to integrate transactions with structural missing value increases drastically the range of application of the proposed framework (see table \ref{ntxs}). Moreover it adds an other perspective to the framework: HMM-based features calculated for a small (resp. big) window-size will characterize the short (resp. long) term history. 
%
%
%
%
%
%
%
%
%

\section*{Conclusion}

In this work, we propose an HMM-based feature engineering strategy that allows us to incorporate sequential knowledge in the transactions in the form of HMM-based features. These HMM-based features enable a non sequential classifier (Random Forest) to use sequential information for the classification.

The multiple perspective property of our HMM-based automated feature engineering strategy gives us the possibility to incorporate a broad spectrum of sequential information. In fact, we model the genuine and fraudulent behaviours of the merchants and the card-holders according to two features: the timing and the amount of the transactions. Moreover, the HMM-based features are created in a supervised way and therefore lower the need of expert knowledge for the creation of the fraud detection system. The terminal perspective is usually not used in credit card fraud detection and is shown in this paper to greatly help the detection for face-to-face and e-commerce transactions.

This extension to \cite{Lucas2019} consolidates the claims already made with additional experiments and evaluations. More precisely:
\begin{itemize}
\item The feature engineering strategy is shown to perform well for e-commerce and face-to-face credit card fraud detection: the results show an increase in the precision-recall AUC of 18.1\% for the face-to-face transactions and 9.3\% for the e-commerce ones. 
\item Then, the feature engineering strategy is shown to be relevant for various types of classifiers (random forest, logistic regression and Adaboost) and robust to hyperparameters choices made for constructing the features.
\item Finally, the structural missing values limitation of the framework is looked at and several solutions are benchmarked.  
\end{itemize}

HMM-based feature engineering strategy is a powerful tool that is shown to present interesting properties for fraud detection. We can imagine building similar HMM-based features in any supervised task that involve a sequential dataset.

To ensure reproducibility, a source code for calculating and evaluating the proposed HMM-based features can be found at \url{https://gitlab.com/Yvan_Lucas/hmm-ccfd}.

As a future work, it would be interesting to combine the predictions of an LSTM with the prediction of some HMM-based features enhanced Random Forest since these classifiers have been shown to not detect the same frauds in face-to-face transactions by \citep{Jurgovsky2018}.

\subsection{Acknowledgement:} The work has been funded partially by the Bavarian Ministry of Economic Affairs, Regional Development and Energy in the project “Internetkompetenzzentrum Ostbayern".



\bibliographystyle{elsarticle-harv} 
\bibliography{sample.bib}

\end{document}